\documentclass[preprint,review,10pt]{elsarticle}

\journal{Pattern Recognition}
\biboptions{sort&compress}

\usepackage[T1]{fontenc}
\usepackage{amsmath,amssymb}
\usepackage{newtxtext,newtxmath}

\usepackage{booktabs}
\usepackage{makecell}
\usepackage{ragged2e}
\usepackage{tabularx}
\usepackage{siunitx}

\usepackage{graphicx}
\usepackage{float}
\usepackage{placeins}

\usepackage{caption}
\DeclareCaptionFont{eightpt}{\fontsize{8pt}{9.5pt}\selectfont}

\usepackage{subcaption}
\usepackage{enumitem}
\setlist[itemize]{label=\textbullet}

\usepackage[hidelinks]{hyperref}

\hypersetup{
  pdftitle={DRIVE: Modeling Skills at the Reasoning and Interaction Levels for Web Agents under Continual Learning},
  pdfauthor={Xirui Liu, Sihang Zhou, Yanning Hou, Rong Zhou, Haoyuan Chen, Maolin He, Siwei Wang, Hao Chen, Jian Huang}
}
\begin{document}

\begin{frontmatter}

\title{DRIVE: Modeling Skills at the Reasoning and Interaction Levels for Web Agents under Continual Learning}

\author[1]{Xirui Liu}
\author[1]{Sihang Zhou\corref{cor1}}
\ead{sihangjoe@gmail.com}

\author[1]{Yanning Hou}
\author[1]{Rong Zhou}
\author[1]{Haoyuan Chen}
\author[1]{Maolin He}
\author[2]{Siwei Wang}
\author[1]{Hao Chen}
\author[1]{Jian Huang}

\cortext[cor1]{Corresponding author.}

\affiliation[1]{
  organization={College of Intelligence Science and Technology, National University of Defense Technology},
  city={Changsha},
  postcode={410073},
  state={Hunan},
  country={China}
}
\affiliation[2]{
  organization={College of Computer Science and Technology, National University of Defense Technology},
  city={Changsha},
  postcode={410073},
  state={Hunan},
  country={China}
}


\begin{abstract}
Web agents require both high-level reasoning (for task decomposition) and low-level interactions (for page elements manipulation) to conduct different tasks. However, these knowledge types differ fundamentally: reasoning knowledge (e.g., booking a flight requires first searching for routes) is abstract and transferable across websites, while interaction knowledge (e.g., clicking the Search button at a specific coordinate on Site A) depends heavily on page-specific contexts. Existing methods store experiences uniformly. This creates a dilemma: abstract representations lose executability on concrete pages, while concrete representations fail to generalize across domains. This entanglement limits capability accumulation: on new websites, agents either fail to recognize reusable task logic due to surface-level differences or attempt infeasible actions from outdated page structures. To disentangle them, we propose DRIVE, a dual-level skill modeling framework separating historical experience into natural language reasoning skills, which capture transferable task logic, and programmatic interaction skills, grounding abstract actions to executable operations. A scene-aware coordination mechanism adaptively retrieves and invokes these dual-level skills based on task semantics. DRIVE also uses skill-level reflection to identify hierarchy-specific failure modes, enabling targeted skill library expansion and refinement. Experiments across five WebArena domains show DRIVE attains an average task success rate of 52.8\%, exceeding the skill-free baseline by 7.3 percentage points. Further ablations show reasoning and interaction skills provide distinct, complementary benefits, supporting separation of transferable task logic from executable page-level operations.
\end{abstract}


\end{frontmatter}



\section{Introduction}
\label{sec:Introduction}
As large language models improve the ability of web agents to complete tasks \cite{zheng2024gpt,he2024webvoyager,lai2024autowebglm}, a central question is how these agents can achieve stable, long-term adaptation in dynamic web environments~\cite{liu2024domain}. Compared with static benchmarks, web tasks present distinct challenges because their states change over time and their solution paths are often diverse \cite{zheng2024webolympus,pan2406webcanvas,ye2025realwebassist,xue2025illusion}. Even when the goal is similar, action sequences may shift as page components and layouts are updated, and the appropriate next action often depends on the current page state and its surrounding context \cite{lee2025learning,gou2024navigating}. As a result, a web agent cannot rely solely on on-the-fly reasoning or task-local exploration, as in many existing agent frameworks \cite{yao2022react,he2024webvoyager,lai2024autowebglm}. It must also accumulate, reuse, and revise experience from past interactions\cite{han2024adaptive}. Such experience may include task-level regularities for decomposing recurring goals, as well as interaction-level cues indicating how particular page states enable or constrain actions. This longer-term use of experience is essential for robust adaptation in changing web environments.

To make web agents more adaptive over long horizons, recent work has studied how to store and reuse past experience. Representative methods distill experience into natural-language reflections or rules that guide later decisions \cite{zhao2024expel}, or retain interaction trajectories and demonstrations that can be consulted when similar tasks arise \cite{wang2024agent,liu2025contextual,zheng2504skillweaver,zhou2025proposer}. Reusing such experience can reduce repeated trial and error. For instance, an agent may learn from earlier failures that a task should be decomposed into search, filtering, and confirmation steps, or that a particular page state requires closing a modal window before the target action becomes available.

More recent work has studied programmatic skills as executable representations of reusable web experience \cite{prabhu2025walt,zhong2026actionengine}. Rather than treating raw trajectories only as demonstrations, these methods abstract recurring interaction patterns into callable functions so that an agent can perform structurally similar operations by providing task- or page-specific arguments \cite{wang2024agent,prabhu2025walt}. This formulation is well suited to interaction-level knowledge, since it represents procedural constraints, action order, and executable grounding more directly than natural-language reflections \cite{zhong2026actionengine}. For example, a form-filling trajectory can be converted into a parameterized function that locates the relevant input fields, fills in the required values, and submits the form. The agent can then reuse this operation without planning each low-level action again from scratch \cite{prabhu2025walt}.

These efforts point to a deeper challenge: different forms of experience tend to encode different levels of knowledge~\cite{liu2025class}. Natural-language reflections can express transferable task-level lessons, such as recognizing when an ineffective strategy should be revised, but they often provide too little grounding to determine which page element should be operated on. Trajectories and programmatic skills, by contrast, retain richer executable details, including clicks, inputs, page states, and procedural constraints. These details, however, usually transfer only when the new task shares similar page layouts, element structures, or interaction contexts. The main limitation is therefore not simply the choice of representation, but the absence of a clear separation and coordination between reasoning knowledge and interaction knowledge. When these two levels are stored and reused as a single experience object, agents face a persistent trade-off between cross-site transferability and page-level executability.

Motivated by this observation, we propose DRIVE, a dual-level skill modeling framework that separates historical web experience into transferable reasoning skills and executable interaction skills. Rather than forcing all experience into a single representation, DRIVE uses a form suited to each level of reuse. Reasoning skills are written in natural language and capture knowledge for task understanding and decision making, while interaction skills are represented as programs that encode executable page-operation patterns and action constraints. To support reuse, DRIVE links each skill to its usage scenario, described by task semantics and page-context conditions. It then uses a scene-aware mechanism to retrieve and coordinate dual-level skills for the current task-page scene. DRIVE also uses skill-level failure feedback to revise, expand, and deduplicate the skill library over time, forming a continual learning pipeline based on layered representation, coordinated reuse, and closed-loop updating.

We evaluate DRIVE on five website domains from WebArena~\cite{zhou2023webarena}. DRIVE achieves an average task success rate of 52.8\%, improving over the skill-free baseline by 7.3 percentage points. Its performance also increases as more historical trajectories are used for skill induction and updating. These results suggest that dual-level skill modeling provides an effective way to accumulate experience, reuse capabilities, and support continual improvement in web agents.

The main contributions of this work are as follows.
\begin{itemize}
\item We motivate DRIVE by identifying the heterogeneity between reasoning knowledge and interaction knowledge in web interaction experience, highlighting the limitation of treating historical experience as a unified representation.
    
    \item We present DRIVE, a dual-level skill modeling framework that represents historical experience as natural-language reasoning skills and programmatic interaction skills. DRIVE reuses and updates these skills over time by retrieving and invoking them according to the current scene, and by incorporating failure feedback at the skill level.
    
    \item We evaluate DRIVE across the five website domains in WebArena. The results show that DRIVE improves task success rates, and its performance continues to increase as experience accumulates.
\end{itemize}

\section{Related work}
\label{sec:Related_work}

Prior work has improved web agents by reusing past trajectories through episodic memory and experience replay~\cite{wang2025continual}. AWM~\cite{wang2024agent} extracts reusable workflows from previous interactions, CER~\cite{liu2025contextual} retrieves relevant patterns from a replay buffer, WebCoach~\cite{liu2025webcoach} compresses navigation histories into reusable guidance, and ExpSeek~\cite{zhang2026expseek} intervenes with prior experience when the agent is uncertain. Together, these methods show that accumulated trajectories can support decisions in later tasks. In most cases, however, the experience is reused as a single object, such as a memory entry or workflow. This design is useful for recalling high-level strategies, but it leaves low-level interaction knowledge largely implicit, including the knowledge needed to carry out those strategies on a concrete web page. For example, a summary may tell an agent to search, filter, and confirm, yet the agent may still fail when it must identify the right interface element or handle page-specific constraints.

Executable skill induction. A related line of work studies how web interaction trajectories can be abstracted into programmatic or callable skills. SkillWeaver~\cite{zheng2504skillweaver} synthesizes reusable web skills as APIs through exploration and practice, and CASCADE~\cite{huang2025cascade} and ALITA~\cite{qiu2025alita} study how repeated interaction patterns can be exposed as callable tools or skills. These methods show that interaction data can be converted into reusable operations rather than kept only as contextual memory. The resulting skills or policies, however, are usually represented as single procedural units. This improves executability, but offers limited support for capturing the reasoning experience that explains when and why an operation should be used. Other self-improving frameworks, including WebRL~\cite{qi2025webrltrainingllmweb}, WebAgent-R1~\cite{wei2025webagent}, and Agent Q~\cite{putta2024agent}, improve agent behavior through online exploration, reinforcement learning, or search-based correction. Their feedback signals are typically used for overall behavior optimization rather than skill-level revision. This leaves existing methods limited in how they accumulate reasoning experience and support the continual refinement of reusable skills.

Our work connects experience reuse with executable skill induction, but argues that both face a shared representational problem: experience derived from trajectories is often treated as a single reusable object, even though it contains knowledge with different levels of abstraction and different grounding requirements. DRIVE addresses this abstraction--grounding mismatch by representing experience at the level where it can be reused most effectively. Instead of treating memory retrieval and executable skills as separate components, DRIVE separates the knowledge that can generalize across tasks from the knowledge that must be grounded in a specific web context. This decomposition provides a stronger basis for continual skill evolution, allowing reusable experience to be refined and expanded according to its level of abstraction rather than updated as one monolithic memory or procedural unit.
\section{Method}
\label{sec:Method}
\begin{figure}[H]
    \centering
    \includegraphics[width=\linewidth,trim=0.2 0cm 0 0cm,clip]{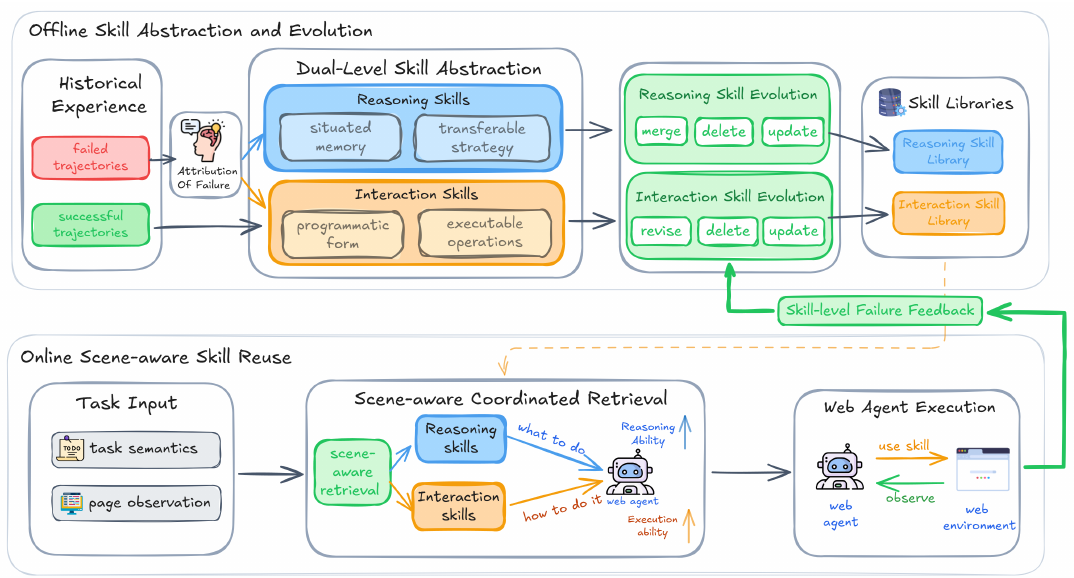}
    \caption{Overview of DRIVE. DRIVE consists of an offline stage for skill abstraction and evolution and an online stage for scene-aware skill reuse. The offline stage abstracts historical web interaction trajectories into reasoning and interaction skills, which are organized in hierarchical skill libraries and refined through evolution. During online execution, the agent retrieves relevant skills according to task semantics and page observations, and feeds execution feedback back to the offline stage to form a closed-loop refinement process.}
    \label{fig:framework}
\end{figure}

Figure~\ref{fig:framework} illustrates the overall framework of DRIVE. DRIVE follows an offline--online loop for building, reusing, and refining web-agent skills. In the offline phase, historical trajectories are abstracted into two skill libraries. Reasoning skills capture reusable task logic, while interaction skills record page-grounded operation experience. DRIVE further analyzes failures at the skill level, so reasoning failures update the reasoning library, whereas interaction failures repair the interaction library. In the online phase, given a task instruction and the current page observation, DRIVE retrieves skills that match both the task intent and the webpage context. The reasoning skill provides corrective guidance to avoid recurring reasoning errors, while the interaction skill reuses successful page-level experience to reduce execution failures. The resulting feedback is returned to the offline phase to refine the libraries. By decoupling transferable reasoning from page-grounded interaction while coordinating them during execution, DRIVE enables web agents to accumulate capabilities across dynamic web environments.

\subsection{Problem Formulation}
\label{sec:problem_formulation}

We study a web agent built on a language model backbone $\mathcal{L}$ that interacts with a web environment to solve a natural-language task instruction $q$. At environment step $t$, let $s_t$ denote the underlying environment state, and let $o_t$ denote the corresponding observation available to the agent. The agent acts in a predefined primitive action space $\mathcal{A}$, and the environment evolves according to
$$s_{t+1} = \mathcal{T}(s_t, a_t),$$
where $a_t \in \mathcal{A}$ is the action executed at step $t$. The interaction continues until the agent emits a termination action or the environment reaches a terminal condition.

To adapt continuously in dynamic web environments, we assume that the agent maintains a dual-level skill library
$$\mathcal{K} = (\mathcal{K}^{r}, \mathcal{K}^{i}),$$
where $\mathcal{K}^{r}$ is a reasoning skill library and $\mathcal{K}^{i}$ is an interaction skill library. We also use $\mathcal{K}^{\mathrm{all}} = \mathcal{K}^{r} \cup \mathcal{K}^{i}$ to denote the union of all skill entries. Reasoning skills capture transferable knowledge for task understanding and decision making, such as subgoal decomposition, strategy selection, state assessment, and outcome verification. Interaction skills capture executable web procedures grounded in page context, including element grounding, action sequences, triggering conditions, and execution constraints.

At each step, the agent retrieves relevant reasoning and interaction skills conditioned on the task instruction and the current observation:
$$(k_t^r, d_{k_t^r}) = \mathrm{Retrieve}^{r}(q, o_t, \mathcal{K}^{r}), \qquad (k_t^i, d_{k_t^i}) = \mathrm{Retrieve}^{i}(q, o_t, \mathcal{K}^{i}).$$
The retrieved reasoning skill provides high-level guidance for decision making, for example by identifying the next subgoal or action intent. The retrieved interaction skill is then invoked directly as an executable procedure to carry out the intended web operation under the current page context. In this way, reasoning skills guide what to do, while interaction skills determine how to do it.

A task execution produces an interaction episode
$$\tau = \left(q, \{(o_t, a_t)\}_{t=1}^{T}, y\right),$$
where $T$ is the trajectory length and $y \in \{0,1\}$ indicates whether the task is completed successfully. Over time, the agent accumulates successful episodes $\Gamma_{\mathrm{succ}}$ and failed episodes $\Gamma_{\mathrm{fail}}$.

Our goal is to continually update the dual-level skill library from historical interaction experience so that the resulting skills can be reused to improve future task completion in changing web environments. Let $n$ denote the skill evolution round, and let $\Phi_n$ denote the skill-level feedback collected before round $n$. The self-evolution process is defined as
$$\left(\mathcal{K}^{r,(n+1)}, \mathcal{K}^{i,(n+1)}\right) = \mathrm{Update}\!\left( \mathcal{K}^{r,(n)}, \mathcal{K}^{i,(n)}, \Gamma_{\mathrm{succ}}, \Gamma_{\mathrm{fail}}, \Phi_n \right),$$
where $\mathrm{Update}(\cdot)$ denotes the skill evolution operator that revises existing skills, induces new skills, and removes obsolete or redundant ones based on accumulated experience. During online deployment, invoking retrieved skills also produces skill-level feedback, especially information about why skill use fails under the current task and page context. This failure feedback is recorded and incorporated into later updates, allowing both reasoning and interaction skills to be refined over time.

\subsection{Skill Representation}
\label{sec:skill_representation}

To support continual self-evolution, the agent maintains a dual-level skill library
$$\mathcal{K} = (\mathcal{K}^{r}, \mathcal{K}^{i}),$$
where $\mathcal{K}^{r}$ and $\mathcal{K}^{i}$ denote the reasoning skill library and the interaction skill library, respectively. We use $\mathcal{K}^{\mathrm{all}} = \mathcal{K}^{r} \cup \mathcal{K}^{i}$ to denote all skill entries. The two types of skills capture complementary aspects of historical experience. Reasoning skills encode transferable knowledge about what to do, whereas interaction skills encode executable knowledge about how to do it. Each skill is also associated with a scenario descriptor that summarizes the task semantics and page-context conditions under which the skill is applicable. This descriptor serves as the interface between offline skill induction and online skill retrieval.

Interaction skills in $\mathcal{K}^{i}$ encode reusable executable web procedures grounded in page context. They are primarily induced from successful trajectories and can be further refined using execution-level failure feedback collected during online deployment. Rather than functioning as a static open-loop mapping, each interaction skill is represented as a parameterized executable procedure (or sub-policy). Invoking an interaction skill $k^{i}$ under a specific web context $C \in \mathcal{C}$ with input arguments $X \in \mathcal{X}$ triggers a dynamic, closed-loop execution process against the environment. This process yields a realized action trace $\alpha$ and a final execution outcome $e$:
$$k^{i}(C, X) \rightarrow (\alpha, e),$$
where $\mathcal{C}$ denotes the space of web contexts (including page structure, available elements, and local state conditions), $\mathcal{X}$ denotes the space of input arguments required for execution, $\alpha \in \mathcal{A}^{*}$ is the finite sequence of primitive actions executed during the invocation, and $e \in \mathcal{Y}$ is the execution outcome. During execution, $k^{i}$ actively interacts with the page (e.g., iteratively verifying locators and handling dynamic UI changes) rather than blindly emitting a predefined sequence. In this sense, an interaction skill is not simply a behavioral hint, but a robust executable module that directly supports action execution on the web page.

Reasoning skills in $\mathcal{K}^{r}$ encode reusable guidance for task understanding and decision making. Unlike interaction skills, they do not execute web actions directly. Instead, they provide high-level support for subgoal decomposition, strategy selection, state assessment, and outcome verification. We represent a reasoning skill as
$$k^{r} = \langle \mathcal{M}, \mathcal{B}, \mathcal{V} \rangle,$$
where $\mathcal{M}$ summarizes a previously observed reasoning error pattern or decision deficiency, $\mathcal{B}$ specifies the corresponding corrected reasoning principle or behavioral guidance, and $\mathcal{V}$ defines explicit verification instructions for checking whether the current decision process satisfies the intended task requirements. During execution, reasoning skills are injected into the model context as structured guidance, helping the agent avoid recurring reasoning errors and make more reliable decisions in related scenarios.

\paragraph{Scenario Descriptor}
Each skill entry $(k,d_k) \in \mathcal{K}^{\mathrm{all}}$ contains a skill $k$ and its associated scenario descriptor
$$d_k = \langle \mathcal{U}_k, \mathcal{W}_k \rangle,$$
where $\mathcal{U}_k$ denotes the applicable web context of skill $k$, such as URL patterns or page-level environment cues, and $\mathcal{W}_k$ denotes the semantic task scenario captured by skill $k$. Given the current instruction $q$ and observation $o_t$, the agent uses $d_k$ to retrieve candidate skills and determine whether a skill should be invoked under the current context.

\subsection{Skill Induction from Successful Trajectories}

A successful episode $\tau \in \Gamma_{\mathrm{succ}}$ contains reusable web operation patterns for accomplishing a task, but it remains tied to a specific task instance and page context. Instead of replaying such instance-specific experience directly, DRIVE abstracts it into a parameterized interaction skill that can be reused across related tasks and similar web scenarios.

Given a successful episode $\tau$, we prompt the language model to synthesize an executable interaction skill $k_{\tau}^{i}$. This process abstracts task-specific entities in the instruction and observation history into input arguments in $\mathcal{X}$, converting the original episode into a reusable procedure conditioned on context. The induced interaction skill encapsulates the necessary closed-loop execution logic and is represented as:
$$k_{\tau}^{i}(C, X) \rightarrow (\alpha, e),$$
where $\mathcal{C}$ denotes the space of web contexts required for execution, $\mathcal{X}$ denotes the space of parameterized task-specific inputs, $\alpha \in \mathcal{A}^{*}$ is the dynamic action trace generated during interaction, and $e \in \mathcal{Y}$ is the final execution outcome.

Along with the executable procedure, the language model also generates a scenario descriptor $d_{k_{\tau}^{i}}$ that summarizes the task semantics and page-context conditions under which the skill applies. This descriptor serves as the retrieval key for later online invocation.

The induced interaction skill and its associated scenario descriptor are then accumulated into a new interaction skill pool for the current round, rather than updating the main library immediately:
$$\mathcal{K}_{\mathrm{new}}^{i} \leftarrow \mathcal{K}_{\mathrm{new}}^{i} \cup \{(k_{\tau}^{i}, d_{k_{\tau}^{i}})\}.$$

\subsection{Counterfactual Failure Attribution from Failed Trajectories}
\label{sec:failed_skill_induction}

Failed episodes provide useful signals for self-evolution, but their underlying causes are often entangled. To identify the source of failure in a failed episode $\tau \in \Gamma_{\mathrm{fail}}$, we introduce an LM-based attribution module
\[
c = f_{\mathrm{attr}}(\tau),
\]
where $c \in \mathcal{E}_{\mathrm{fail}} = \{\mathrm{err}^{i}, \mathrm{err}^{r}\}$ denotes the inferred failure type. Instead of treating each failed episode as a uniform negative example, we prompt the language model to perform a counterfactual analysis of the error trace by asking whether the task could have been completed if the agent had preserved its high-level intent but followed a different interaction procedure. If the answer is yes, the failure is attributed to an interaction-level error ($c = \mathrm{err}^{i}$); otherwise, it is attributed to a reasoning-level error ($c = \mathrm{err}^{r}$).

Based on the attributed failure type, the failed episode is converted into a corresponding skill entry:
\[
G_{\mathrm{fail}}(\tau, c)=
\begin{cases}
(k_{\tau}^{i}, d_{k_{\tau}^{i}}), & c = \mathrm{err}^{i},\\
(k_{\tau}^{r}, d_{k_{\tau}^{r}}), & c = \mathrm{err}^{r},
\end{cases}
\]
where $G_{\mathrm{fail}}(\cdot)$ maps a failed episode to either an interaction-skill entry or a reasoning-skill entry.

A reasoning-level error indicates that the failure comes from incorrect task understanding, flawed decomposition, or invalid assumptions about the environment, rather than from the execution procedure itself. In this case, the failed episode is converted into a reasoning skill
\[
k_{\tau}^{r} = \langle \mathcal{M}, \mathcal{B}, \mathcal{V} \rangle,
\]
where $\mathcal{M}$ summarizes the observed reasoning error pattern, $\mathcal{B}$ specifies the corrected reasoning principle or behavioral guidance, and $\mathcal{V}$ defines explicit verification instructions. These skills serve as structured corrective guidance and are injected into the model context in future related tasks to reduce similar reasoning failures. The induced reasoning skill and its associated scenario descriptor are then accumulated into a new reasoning skill pool via
\[
\mathcal{K}_{\mathrm{new}}^{r} \leftarrow \mathcal{K}_{\mathrm{new}}^{r} \cup \{(k_{\tau}^{r}, d_{k_{\tau}^{r}})\}.
\]

By contrast, an interaction-level error indicates that the agent follows an appropriate high-level strategy but fails during web interaction because of page-specific interruptions, missing elements, selector mismatch, or brittle interaction paths. In this case, we synthesize an interaction skill $k_{\tau}^{i}$ that encodes a corrected executable procedure for the failed task pattern. Compared with interaction skills induced from successful episodes, these skills may also include recovery logic, such as condition checking, exception handling, or conditional retries, to improve robustness under similar web conditions. The induced interaction skill and its associated scenario descriptor are then added to the new interaction skill pool via
\[
\mathcal{K}_{\mathrm{new}}^{i} \leftarrow \mathcal{K}_{\mathrm{new}}^{i} \cup \{(k_{\tau}^{i}, d_{k_{\tau}^{i}})\}.
\]

For both failure types, the newly induced skill is further associated with a scenario descriptor, denoted by $d_{k_{\tau}^{r}}$ or $d_{k_{\tau}^{i}}$, which supports future retrieval and conditional invocation under similar task semantics and page-context conditions. In this way, failed episodes are not discarded as unsuccessful experience, but are explicitly transformed into reusable knowledge that supports continual self-evolution.

\subsection{Scenario-Aware Skill Retrieval and Invocation}
\label{sec:skill_invocation}

To reuse accumulated experience in a new task, the agent performs scenario-aware retrieval and coordinated invocation over the dual-level skill library
$$\mathcal{K}=(\mathcal{K}^r,\mathcal{K}^i).$$
At environment step $t$, the current observation $o_t$ provides the full environmental signals. From this, we extract the local page context, including the active URL $u_t$ and the current interface state, denoted by
$$C_t = \mathrm{ExtractContext}(o_t).$$
Each skill entry $(k,d_k) \in \mathcal{K}^{\mathrm{all}}$ contains a scenario descriptor
$$d_k=\langle \mathcal{U}_k,\mathcal{W}_k\rangle,$$
where $\mathcal{U}_k$ characterizes the page-context conditions under which skill $k$ applies, and $\mathcal{W}_k$ summarizes the task-semantic conditions under which skill $k$ is relevant.

Given the current task instruction $q$ and observation $o_t$, we use a two-stage mechanism for both reasoning and interaction skills. We first perform structural filtering by retaining only the skills whose applicable page context is compatible with the current webpage state, yielding candidate sets
$$\mathcal{K}^{r}_{\mathrm{cand},t} = \{ (k,d_k) \in \mathcal{K}^r \mid \mathrm{Match}(C_t,\mathcal{U}_k)=1 \},$$
$$\mathcal{K}^{i}_{\mathrm{cand},t} = \{ (k,d_k) \in \mathcal{K}^i \mid \mathrm{Match}(C_t,\mathcal{U}_k)=1 \}.$$
We then derive a compact semantic representation of the current task context from $q$ and $o_t$ and compare it with the task-semantic descriptors $\mathcal{W}_k$ associated with the candidate skills. This descriptor-level matching step ranks candidate skills without requiring the full contents of all skills to be injected into the model context.

To avoid conflicts among multiple applicable skills, DRIVE treats the retrieved skills as candidates rather than actions to be executed jointly. At each step, the agent applies a singleton activation policy within each skill level: it selects at most one reasoning skill and at most one interaction skill according to their compatibility with the current task-page scene. Formally,
$$(k_t^r, d_{k_t^r})=\mathrm{Select}^r(q,o_t,\mathcal{K}_{\mathrm{cand},t}^r), \qquad
(k_t^i, d_{k_t^i})=\mathrm{Select}^i(q,o_t,\mathcal{K}_{\mathrm{cand},t}^i),$$
where
$$\mathrm{Select}^{\ell}(q,o_t,\mathcal{K}_{\mathrm{cand},t}^{\ell})
\in \mathcal{K}_{\mathrm{cand},t}^{\ell} \cup \{\varnothing\}, \qquad \ell \in \{r,i\}.$$
The selection is based on the match between the current task semantics, page context, and the scenario descriptor $d_k$. If no candidate is sufficiently compatible with the current scene, the selector returns $\varnothing$, and the agent falls back to primitive action generation. Thus, multiple skills may be retrieved as candidates, but they are not invoked simultaneously.

When a reasoning skill is selected, it has the form
$$k_t^r = \langle \mathcal{M}_t, \mathcal{B}_t, \mathcal{V}_t \rangle$$
and is injected into the model context as structured guidance for decision making. Specifically, it reminds the agent of the relevant reasoning error pattern $\mathcal{M}_t$, provides the corresponding corrected strategy $\mathcal{B}_t$, and highlights the verification instructions $\mathcal{V}_t$ to guide subsequent reasoning and action planning.

When an interaction skill $k_t^i$ is selected, the agent first checks whether its scenario descriptor is consistent with the current reasoning intent and page context. If the selected interaction skill is compatible, the agent instantiates its required input arguments $X_t$ from the current task and observation, and then executes the skill under the current web context:
$$(\alpha_t, e_t) = k_t^i(C_t, X_t), \qquad \alpha_t \in \mathcal{A}^{*}, \ e_t \in \mathcal{Y},$$
where $C_t$ denotes the current page context, $X_t$ denotes the instantiated task-specific arguments, $\alpha_t$ denotes the realized finite sequence of primitive actions executed during the invocation, and $e_t$ denotes the corresponding execution outcome. If the selected interaction skill is incompatible or fails its applicability check, the agent does not invoke it and instead falls back to primitive action generation.

Compared with token-level generation of primitive actions, this invocation mode directly reuses a context-conditioned executable interaction procedure for the current task pattern and returns both the action trace and the local execution result. In this way, scenario-aware retrieval enables the agent to use the complementary roles of the two skill types together: reasoning skills guide what to do, while interaction skills support how to do it under the current page context.

\subsection{Skill Management and Continual Refinement}
\label{sec:skill_management}

As experience accumulates, the dual-level skill library
$$\mathcal{K}=(\mathcal{K}^r,\mathcal{K}^i)$$
continues to grow. DRIVE manages the library at each evolution round to remove invalid interaction skills and redundant reasoning skills. Since the two skill types have different feedback signals, they are updated with different operators.

\paragraph{Skill-level feedback}
When a skill is invoked, DRIVE records its execution feedback as
$$\phi = \left\langle k_\phi, d_{k_\phi}, q_\phi, C_{t,\phi}, X_{t,\phi}, e_{t,\phi}, \Delta_{t,\phi}, y_\phi \right\rangle$$
where $k_\phi$ is the invoked skill, $d_{k_\phi}$ is its descriptor, $q_\phi$ is the task instruction, $C_{t,\phi}$ is the current page context, $X_{t,\phi}$ is the instantiated input, $e_{t,\phi}$ is the skill execution outcome, $y_\phi$ is the final task label, and $\Delta_{t,\phi}$ is the execution log. The feedback collected before round $n$ is denoted by
$$\Phi_n=\{\phi_1,\phi_2,\ldots,\phi_m\}.$$
For interaction skills, $\Delta_{t,\phi}$ records local signals, including selector matching, URL or page-state changes, returned results, and execution exceptions. For reasoning skills, no clear local error is usually available, so DRIVE mainly uses the final task outcome as a weak utility signal.

\paragraph{Interaction skill repair}
Interaction skills in $\mathcal{K}^i$ are executable procedures ($k^{i} : \mathcal{C} \times \mathcal{X} \rightarrow \mathcal{A}^{*} \times \mathcal{Y}$). They may fail when the UI changes, selectors become invalid, or a modal blocks the intended action. To reduce this brittleness, DRIVE does not bind each operation to a single selector. Instead, the internal executable procedure of each interaction skill is parameterized as a sequence of operation templates:
$$\Theta(k^i) = \left[ (r_1,\Sigma_1), (r_2,\Sigma_2), \ldots, (r_P,\Sigma_P) \right],$$
where $r_j$ is the intended operation and
$$\Sigma_j=(\sigma_{j,1},\sigma_{j,2},\ldots,\sigma_{j,L_j})$$
is an ordered sequence of selectors for grounding its target element. These selectors may come from successful executions on similar pages and may include CSS selectors, XPath, DOM identifiers, text-based locators, accessibility attributes, or nearby-label locators. During execution, DRIVE tries the selectors in $\Sigma_j$ in order and uses the first valid one.

DRIVE treats a case as interaction-skill failure when the skill is retrieved and invoked, but its local effect is missing. This includes four common cases: no selector in $\Sigma_j$ matches the target element; the action is executed but the URL or page state does not change as expected; no expected result is returned; or the final page state violates the skill check. In contrast, if the skill is not retrieved due to descriptor mismatch, the case is not treated as interaction-skill failure.

Given interaction feedback $\Phi_n^i \subseteq \Phi_n$, DRIVE repairs interaction skills by local patching:
$${\Theta(k^i)}^{+} = \mathrm{Patch}^{i}(\Theta(k^i),\Phi_n^i).$$
The patch operator mainly rewrites selector sets rather than regenerating the whole skill. For an operation $r_j$, the updated selector set is
$$\Sigma_j^{+} = \mathrm{UpdateSelector} \left( \Sigma_j, C_{t,\phi}, r_j, \Delta_{t,\phi} \right).$$
A failed selector is demoted or removed. A working selector found in later traces is added to the set. If needed, DRIVE also adds a small fallback branch, such as retrying with another locator or closing an unexpected modal. When a skill still fails after repeated selector-level patches, DRIVE removes it from $\mathcal{K}^i$.

\paragraph{Reasoning skill consolidation}
Reasoning skills in $\mathcal{K}^r$ are natural-language rules for task decisions. They do not directly execute actions, so their errors are harder to localize. DRIVE therefore does not rewrite a reasoning skill after a single failure. Instead, it manages reasoning skills through addition, merging, and deletion.

For each reasoning skill $k^r$, DRIVE records its usage count
$$N(k^r)=\sum_{\phi \in \Phi_n}\mathbf{1}[k_{\phi}=k^r],$$
and its successful usage count
$$S(k^r)=\sum_{\phi \in \Phi_n}\mathbf{1}[k_{\phi}=k^r]\cdot y_{\phi}.$$
The smoothed utility score is
$$\rho(k^r)=\frac{S(k^r)+\lambda}{N(k^r)+2\lambda},$$
where $\lambda$ is a smoothing constant. Since task success may also depend on interaction skills, $\rho(k^r)$ is only a weak signal. DRIVE uses it to identify reasoning skills that are rarely useful.

When several reasoning skills describe similar scenarios and give similar guidance, DRIVE merges them:
$$(\bar{k}^r,d_{\bar{k}^r}) = \mathrm{Merge}^{r}(\mathcal{G}^r),$$
where
$$\mathcal{G}^r= \{(k_1^r,d_{k_1^r}),\ldots,(k_l^r,d_{k_l^r})\}$$
is a group of similar reasoning skills. The merged skill keeps the shared mistake pattern, correction rule, and verification condition, while removing repeated or overly specific content. Reasoning skills with low utility, low usage, or strong overlap with a better skill are deleted. New reasoning-level failures identified by $f_{\mathrm{attr}}$ are added as new skills.

\paragraph{Library update}
Let $\mathcal{K}_{\mathrm{new}}^{i}$ and $\mathcal{K}_{\mathrm{new}}^{r}$ denote the pools of new skills induced from successful and failed trajectories during the current round. At the end of round $n$, DRIVE updates the interaction library by applying a batch patching operator across applicable skills and pruning the invalid ones:
$$\mathcal{K}^{i,(n+1)} = \mathrm{Prune}^{i} \left( \mathrm{BatchPatch}^{i} \left( \mathcal{K}^{i,(n)} \cup \mathcal{K}_{\mathrm{new}}^{i}, \Phi_n^i \right), \Phi_n^i \right).$$
It updates the reasoning library by applying a batch merging operator to groups of similar skills and pruning those with low utility scores based on accumulated feedback:
$$\mathcal{K}^{r,(n+1)} = \mathrm{Prune}^{r} \left( \mathrm{BatchMerge}^{r} \left( \mathcal{K}^{r,(n)} \cup \mathcal{K}_{\mathrm{new}}^{r} \right), \Phi_n^r \right).$$
Thus, interaction-skill management focuses on selector repair and small execution patches, while reasoning-skill management focuses on adding, merging, and deleting high-level rules. This keeps the library compact while allowing it to adapt over time.

\section{Experiments}
\label{sec:experiments}

\subsection{Experimental Setup}
\label{sec:Experiment_Setup}

\subsubsection{Environment and Metrics}
\label{sec:task_and_metrics}
We evaluate DRIVE on WebArena~\cite{zhou2023webarena}, a high-fidelity benchmark with 812 multi-step web tasks spanning five domains:Shopping, CMS, Forum, Gitlab, Map. WebArena is well suited to our setting because its tasks require both long-horizon reasoning and accurate page-level grounding. Following the standard benchmark protocol, we use the automated functional-correctness evaluator and report success rate as the primary metric.

\subsubsection{Baselines}
\label{sec:Baselines}
We compare DRIVE with a standard prompting agent and several recent web-agent frameworks. The standard prompting baseline is the vanilla WebArena agent~\cite{zhou2023webarena}, which maps observations directly to actions without retaining past experience. For observation/action alignment, we include AgentOccam~\cite{yang2024agentoccam}, a strong baseline that improves zero-shot performance through LLM-friendly textual representations and standardized action spaces. For data-centric exploration, we compare with Go-Browse~\cite{gandhi2025go}, which uses offline structured exploration to collect interaction trajectories for policy refinement. We also include two skill-based or hierarchical methods: SteP~\cite{sodhi2023step}, which uses a stack of LLM policies for dynamic control flow, and SkillWeaver~\cite{zheng2504skillweaver}, which discovers and distills experience into executable APIs.

Implementation details. To ensure a fair comparison, all experiments, including DRIVE and the reproduced baselines, use GPT-4.1 as the backbone and are evaluated under the same environment configuration.

\subsubsection{Experiment details}
\label{sec:Experiment details}
We limit the web agent to 20 environment steps per episode. For dataset construction and evaluation, we first manually group tasks on each website by task type. We then use a within-website stratified split, assigning 30\% of tasks to training and 70\% to testing, while keeping the task-type coverage of both partitions as broad as possible. This task-type-aware stratification is designed to test generalization over a diverse and representative task distribution and to reduce bias from skewed task compositions.

Unless otherwise specified, our main agent uses GPT-4.1 as the backbone language model with a decoding temperature of 0.1. For fair comparison, all baseline methods use the same model and decoding configuration whenever applicable, and all methods are evaluated on the same test tasks, with identical task IDs and evaluation protocol.
\subsection{Main Results}
\label{sec:Main_results}

We evaluate DRIVE on WebArena and report success rates across five website domains in Table~\ref{tab:webarena_domain_sr}. DRIVE achieves the best overall performance, with an average success rate of 52.8\%, and consistently outperforms all baselines on Shopping, CMS, Forum, Gitlab, and Map.

Compared with AgentOccam, the strongest baseline in our evaluation, DRIVE improves the average success rate from 45.5\% to 52.8\%, a gain of 7.3 percentage points and a relative improvement of 16.0\%. DRIVE also outperforms Go-Browse and SkillWeaver by 10.6 and 23.1 percentage points, respectively. These results show the value of dual-level skill modeling and continual skill refinement for web agents in dynamic environments.

\begin{table}[t]
\centering
\caption{Performance comparison on the WebArena benchmark. We report task success rates (\%) across five diverse website domains. The best performance is marked in \textbf{bold} and the second-best is \underline{underlined}. Our proposed DRIVE framework consistently outperforms all baseline models, achieving the highest average success rate.}
\label{tab:webarena_domain_sr}
\begin{tabular}{lcccccc}
\toprule
Agent & Shopping & CMS & Forum & Gitlab & Map & Average \\
\midrule
WebArena    & 21.4 & 15.4 &  8.1 & 14.1 & 12.7 & 14.3 \\
SteP        & 37.0 & 24.0 & 59.0 & 32.0 & 30.0 & 36.4 \\
Go-Browse   & \underline{41.2} & 38.4 & 51.6 & 41.2 & 38.4 & 42.2 \\
SkillWeaver & 25.1 & 23.8 & 41.3 & 32.0 & 26.2 & 29.7 \\
AgentOccam  & 37.2 & \underline{41.5} & \underline{60.4} & \underline{41.4} & \underline{46.8} & \underline{45.5} \\
Ours        & \textbf{45.9} & \textbf{46.1} & \textbf{66.7} & \textbf{50.0} & \textbf{55.4} & \textbf{52.8} \\
\bottomrule
\end{tabular}
\end{table}

We also include SteP~\cite{sodhi2023step}, which achieves an average success rate of 36.4\%. However, SteP relies on manually designed, website-specific workflows, which require additional effort to transfer across websites and maintain as the environment changes. In contrast, DRIVE automatically induces dual-level skills from historical trajectories and continually refines the skill library using skill-level feedback, without relying on handcrafted domain-specific policies.
\subsection{Self-Evolution with Increasing Training Experience}
\label{sec:self_evolution_with_increasing_training_experience}

To evaluate whether DRIVE continues to benefit from additional experience, we progressively increase the number of training trajectories used for skill induction and refinement while keeping the test set fixed. This experiment examines how the dual-level skill library evolves as more interaction experience becomes available.

\begin{figure}[h]
    \centering
    \includegraphics[width=0.98\linewidth,trim={2pt 1pt 1pt 0pt},clip]{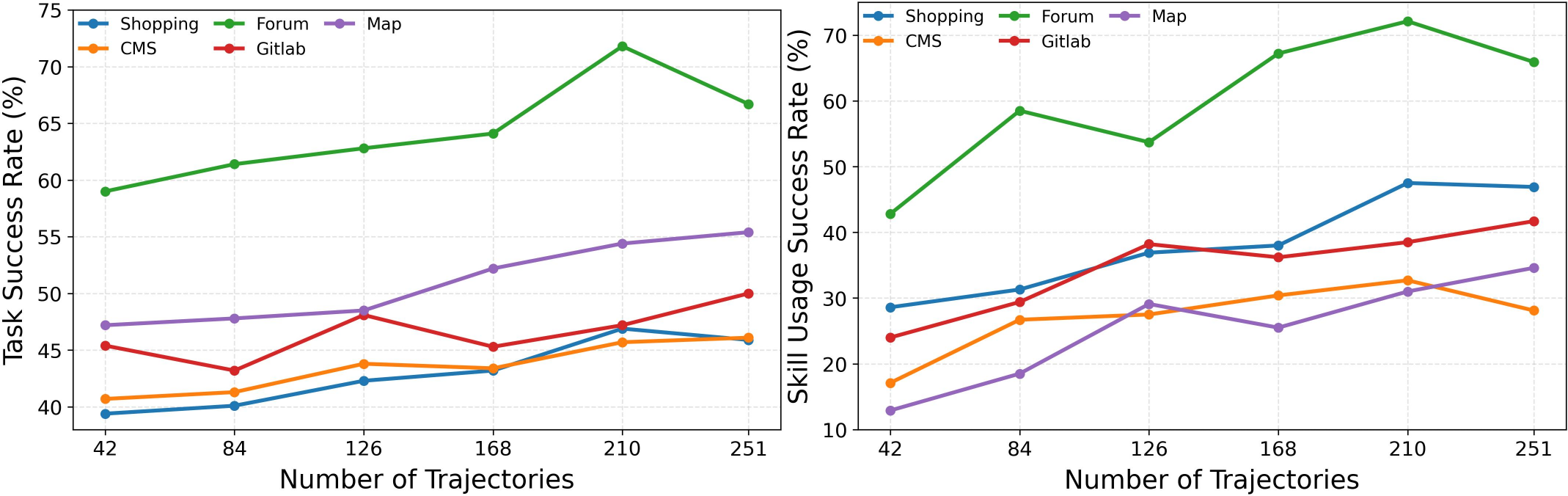}
    \caption{Continuous capability accumulation in DRIVE. As the number of training trajectories increases from 42 to 251, the framework demonstrates steady self-evolution across five WebArena domains. The concurrent improvement in overall task success rate (left) and interaction skill usage success rate (right) validates that DRIVE effectively translates accumulated experience into robust, reusable capabilities.}
    \label{fig:self-evolution-results}
\end{figure}

Figure~\ref{fig:self-evolution-results} shows that DRIVE benefits consistently from additional training experience. As the number of training trajectories increases from 42 to 251, both task success rate and interaction-skill invocation success improve across most WebArena domains, indicating that self-evolution can steadily transform accumulated trajectories into reusable capabilities.

The left panel shows that task success improves most clearly on Shopping, GitLab, and Map as more training trajectories are incorporated. These domains involve relatively complex workflows and substantial interface variation, and therefore benefit more from the joint improvement of interaction skills and reasoning skills. Interaction skills encode reusable executable procedures, while reasoning skills support task understanding, planning, and verification. By contrast, Forum starts from a comparatively strong level and exhibits smaller marginal gains at higher trajectory budgets, suggesting that its dominant workflows are captured relatively early and that later updates mainly improve long-tail cases.

The right panel further shows that the invocation success rate of interaction skills generally increases with additional trajectories, suggesting that self-evolution improves not only skill coverage but also execution reliability. The upward trend is especially clear on Shopping, CMS, Map, and GitLab, where successful execution is more sensitive to page layout variation and page-specific interaction details. Forum maintains the highest invocation success rate throughout, indicating that its recurrent workflows are easier to consolidate into stable reusable interaction skills. Although moderate fluctuations remain at intermediate trajectory budgets, the overall trend suggests that the interaction skill library becomes progressively more robust through repeated failure-driven refinement.

\subsection{Ablation Study}
\label{sec:ablation_study}

\begin{table}[h]
\centering
\caption{Ablation study of DRIVE on the WebArena benchmark. ``Succ.-Int.'' denotes interaction skills distilled from successful trajectories. ``Fail.-Int.'' and ``Fail.-Reas.'' represent interaction and reasoning skills refined from failed trajectories, specifically designed to address operation errors and decision errors, respectively. The full dual-level framework provides the strongest performance.}
\label{tab:ablation-webarena}
\footnotesize
\setlength{\tabcolsep}{7pt}
\renewcommand{\arraystretch}{1.05}
\begin{tabular}{@{}lccccccc@{}}
\toprule
Method &
\shortstack{Shop.\\(\%)} &
\shortstack{CMS\\(\%)} &
\shortstack{Forum\\(\%)} &
\shortstack{GitLab\\(\%)} &
\shortstack{Map\\(\%)} &
\shortstack{Avg.\\(\%)} &
\shortstack{$\Delta$ Avg.\\(pp)} \\
\midrule
Baseline      & 37.2 & 41.5 & 60.4 & 41.4 & 46.8 & 45.5 & -- \\
+ Succ.-Int.  & 41.3 & 44.9 & 65.0 & 48.2 & 49.5 & 49.8 & +4.3 \\
+ Fail.-Int.  & 39.7 & 42.3 & 63.4 & 45.5 & 51.8 & 48.5 & +3.0 \\
+ Fail.-Reas. & 40.1 & 42.8 & 63.1 & 42.6 & 50.1 & 47.8 & +2.3 \\
\textbf{DRIVE} & \textbf{45.9} & \textbf{46.1} & \textbf{66.7} & \textbf{50.0} & \textbf{55.4} & \textbf{52.8} & \textbf{+7.3} \\
\bottomrule
\end{tabular}
\end{table}

Table~\ref{tab:ablation-webarena} details the performance contributions of different skill induction pathways in DRIVE. The baseline agent, devoid of any skill library, achieves an average success rate of 45.5\%. Introducing interaction skills distilled solely from successful trajectories (Succ.-Int.) provides a solid initial gain, raising the success rate to 49.8\% (+4.3 pp). Crucially, the remaining ablations validate our failure-driven reflection mechanism: incorporating interaction skills and reasoning skills refined from failed trajectories yields 48.5\% (+3.0 pp) and 47.8\% (+2.3 pp), respectively. The complete DRIVE framework integrates all induction pathways, achieving the highest performance of 52.8\% (+7.3 pp). These macro-level gains indicate that extracting knowledge from both successful executions and historical failures is indispensable for robust performance.

\subsection{Validation of Failure Attribution}
\label{sec:attribution_validation}

DRIVE abstracts interaction procedures from successful executions, but its skill evolution also relies on learning corrective information from failed trajectories. This requires deciding whether a failure stems primarily from reasoning or from interaction. Because incorrect attribution can introduce poorly aligned knowledge into the skill library, we examine both the reliability of the LM-based attribution module and the effect of attribution quality on downstream performance.

We randomly sample 30 tasks from each of the five WebArena domains, yielding a 150-task evaluation subset. On this subset, the baseline agent achieves a 46.0\% success rate and produces 81 failed trajectories. Two human annotators independently reviewed the execution traces and labeled each failure as either a reasoning error or an interaction error. Disagreements were resolved through discussion. We then compared the LM-based predictions with these human-adjudicated labels.

\begin{figure}[t]
\centering

\begin{minipage}[c]{0.52\linewidth}
    \centering
    \includegraphics[width=\linewidth]{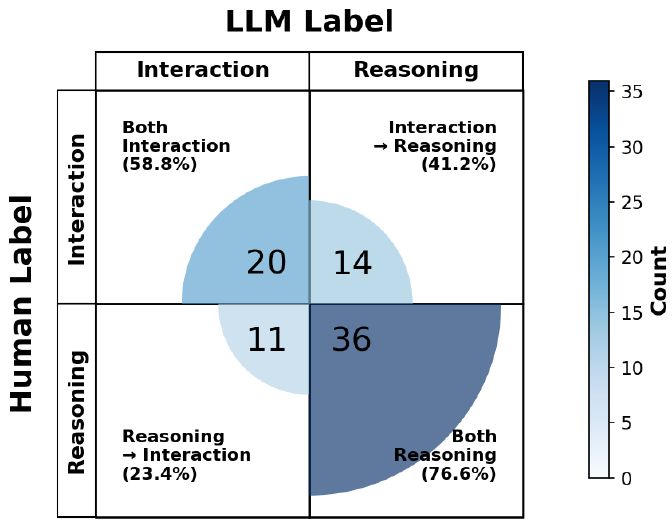}
\end{minipage}
\hfill
\begin{minipage}[c]{0.44\linewidth}
    \centering
    \small
    \setlength{\tabcolsep}{8pt}
    \renewcommand{\arraystretch}{1.05}
    \begin{tabular}{@{}lcc@{}}
    \toprule
    Setting &
    \shortstack{Success\\Rate (\%)} &
    \shortstack{$\Delta$\\(pp)} \\
    \midrule
    No skill       & 46.0 & --   \\
    Human attr.    & \textbf{54.0} & \textbf{+8.0} \\
    LM attr.       & \underline{49.3} & \underline{+3.3} \\
    Random attr.   & 46.7 & +0.7 \\
    Reversed attr. & 44.7 & -1.3 \\
    \bottomrule
    \end{tabular}
\end{minipage}

\vspace{2mm}

\begin{minipage}[t]{0.52\linewidth}
    \centering
    {\footnotesize (a) Attribution confusion matrix}
\end{minipage}
\hfill
\begin{minipage}[t]{0.44\linewidth}
    \centering
    {\footnotesize (b) Effect of attribution quality}
\end{minipage}

\caption{Validation of failure attribution.
(a) Confusion matrix between human-adjudicated and LM-based failure labels, where the LM-based attribution module achieves 69.1\% overall accuracy.
(b) Success rates under different attribution settings on the 150-task subset. Human attribution performs best, while reversed attribution performs worse than the no-skill baseline.}
\label{fig:attribution-validation}
\end{figure}

As shown in Figure~\ref{fig:attribution-validation}(a), the LM-based module achieves 69.1\% overall accuracy against human annotations. It correctly identifies 36 of 47 reasoning errors, corresponding to 76.6\% recall, and 20 of 34 interaction errors, corresponding to 58.8\% recall. These results suggest that the module captures useful failure-type signals, especially for reasoning errors. At the same time, the confusion matrix shows that failure disentanglement remains difficult. In particular, 14 interaction errors are misclassified as reasoning failures. This often happens in cascading failures, where a UI constraint, such as an unhandled pop-up, causes the agent to try implausible alternative actions that the LM interprets as a reasoning error. Conversely, the 11 reasoning errors misclassified as interaction errors often involve premature task termination: the agent incorrectly assumes that the task has been completed and stops, while the LM treats the failure as a localized execution problem.

To assess how attribution quality affects downstream skill evolution, we compare five attribution settings on the 150-task subset. The No skill setting does not use skills derived from failures. Human attribution routes failures according to the human-adjudicated labels, whereas LM attribution uses the default automated module. Random attribution assigns each failed trajectory to a category at random. Reversed attribution deliberately flips the LM-predicted labels, sending predicted reasoning errors to the interaction library and predicted interaction errors to the reasoning library.

Figure~\ref{fig:attribution-validation}(b) shows the effect of attribution quality on task performance. Human attribution achieves the highest success rate, 54.0\%, which is 8.0 percentage points above the baseline. This result indicates that failed trajectories are most useful when they are routed to the appropriate skill type. LM attribution reaches 49.3\%, a gain of 3.3 percentage points, and outperforms both the no-skill baseline and random attribution, which improves by only 0.7 percentage points. In contrast, reversed attribution reduces the success rate to 44.7\%, or 1.3 percentage points below the baseline. This drop supports the separation of skill representations: placing abstract reasoning failures into programmatic interaction skills, or turning UI-specific failures into general natural-language reasoning guidance, contaminates the corresponding skill libraries and weakens later task execution. Although LM-based attribution is still an imperfect approximation of human judgment, it offers a practical mechanism for supporting skill evolution at both levels.

\subsection{Error-Type Analysis of Dual-Level Skills}
\label{sec:error_type_analysis}

Beyond the overall ablation results, we conduct a fine-grained error analysis to examine how the two types of skills affect different failure modes. We use the same 100 randomly sampled test tasks for all skill configurations. Each configuration is evaluated three times under the same evaluation protocol, and we report the mean and standard deviation across runs.

We define a fixed failure taxonomy, shown in Table~\ref{tab:error_taxonomy}, that distinguishes \textbf{reasoning errors} from \textbf{interaction errors}. Reasoning errors reflect failures in task interpretation, exploration, or decision making, whereas interaction errors reflect failures in executing the intended web operation. Successful trajectories are counted separately. Each failed trajectory is assigned to exactly one error category, so the three proportions in Table~\ref{tab:error_type_analysis} are normalized over the same task set and sum to 100\% up to rounding. GPT-4.1 is used only as a post-hoc classifier: given the task instruction, execution trace, and final outcome, it maps each failed trajectory to the fixed human-designed taxonomy.

\begin{table}[t]
\centering
\caption{Failure taxonomy used for error-type analysis. Each failed trajectory is assigned to one of the two mutually exclusive categories.}
\label{tab:error_taxonomy}
\footnotesize
\renewcommand{\arraystretch}{1.15}
\setlength{\tabcolsep}{5pt}
\begin{tabularx}{0.95\linewidth}{
@{}
>{\RaggedRight\arraybackslash}p{0.26\linewidth}
>{\RaggedRight\arraybackslash}X
@{}}
\toprule
\textbf{Error Category} & \textbf{Specific Failure Modes} \\
\midrule
Reasoning Errors
& Answer omission; insufficient exploration; task misunderstanding;
false no-data judgment; empty answer. \\

Interaction Errors
& Repeated execution; multi-field form-filling failure; oscillatory
interaction loop; repeated backtracking. \\
\bottomrule
\end{tabularx}
\end{table}

\begin{table}[t]
\centering
\caption{Error-type analysis on 100 randomly sampled test tasks. Results are reported as mean $\pm$ standard deviation over three runs. Correct tasks and the two error categories are normalized over the same task set.}
\label{tab:error_type_analysis}
\footnotesize
\renewcommand{\arraystretch}{1.12}
\setlength{\tabcolsep}{3.5pt}
\begin{tabularx}{\linewidth}{
@{}
>{\RaggedRight\arraybackslash}p{0.34\linewidth}
>{\centering\arraybackslash}X
>{\centering\arraybackslash}X
>{\centering\arraybackslash}X
@{}}
\toprule
\textbf{Method}
& \makecell{\textbf{Correct}\\\textbf{tasks (\%)}}
& \makecell{\textbf{Reasoning}\\\textbf{errors (\%)}}
& \makecell{\textbf{Interaction}\\\textbf{errors (\%)}} \\
\midrule
No skill
& $42.3 \pm 3.06$
& $35.7 \pm 2.52$
& $22.0 \pm 3.57$ \\

Reasoning-only skills
& $48.0 \pm 1.81$
& $30.3 \pm 2.36$
& $21.7 \pm 2.08$ \\

Interaction-only skills
& $47.7 \pm 2.52$
& $33.3 \pm 0.58$
& $19.0 \pm 2.17$ \\

Reasoning + Interaction skills
& $\mathbf{52.3} \pm 2.52$
& $\mathbf{30.0} \pm 2.03$
& $\mathbf{17.7} \pm 1.53$ \\
\bottomrule
\end{tabularx}
\end{table}

Table~\ref{tab:error_type_analysis} shows that the two skill types affect different parts of the failure distribution. Reasoning-only skills mainly reduce reasoning errors, while leaving the interaction-error rate close to that of the no-skill setting. In contrast, interaction-only skills primarily reduce interaction errors, but do not provide the same reduction in reasoning failures. This pattern is consistent with the intended separation between task-level decision support and page-level execution support.

The full DRIVE configuration combines these two effects. On the sampled subset, it achieves the highest correctness rate while maintaining a reasoning-error rate comparable to the reasoning-only setting and further reducing interaction errors. These results provide evidence that the two skill levels are complementary: reasoning skills help the agent avoid failures in task understanding, exploration, and stopping-condition judgment, whereas interaction skills improve the reliability of executable operations under concrete page constraints.
\subsection{Case Analysis}

We present two representative cases to illustrate why web agents require separate modeling of reasoning skills and interaction skills. In web tasks, failures do not all call for the same kind of correction. Some errors stem from high-level task understanding and stopping-condition judgment, and therefore require reasoning-level guidance. Others arise from brittle execution on dynamic interfaces and therefore require reliable executable procedures. A single skill form does not handle both cases well. Natural-language guidance is often not enough for complex page interactions, while programmatic skills are not suitable for high-level strategy and intent judgment. For this reason, reasoning skills and interaction skills should be modeled separately and invoked according to the demands of the current task.

\begin{figure}[h]
    \centering
    \includegraphics[width=\linewidth,trim=0.1cm 0.1cm 0.1cm 0.1cm,clip]{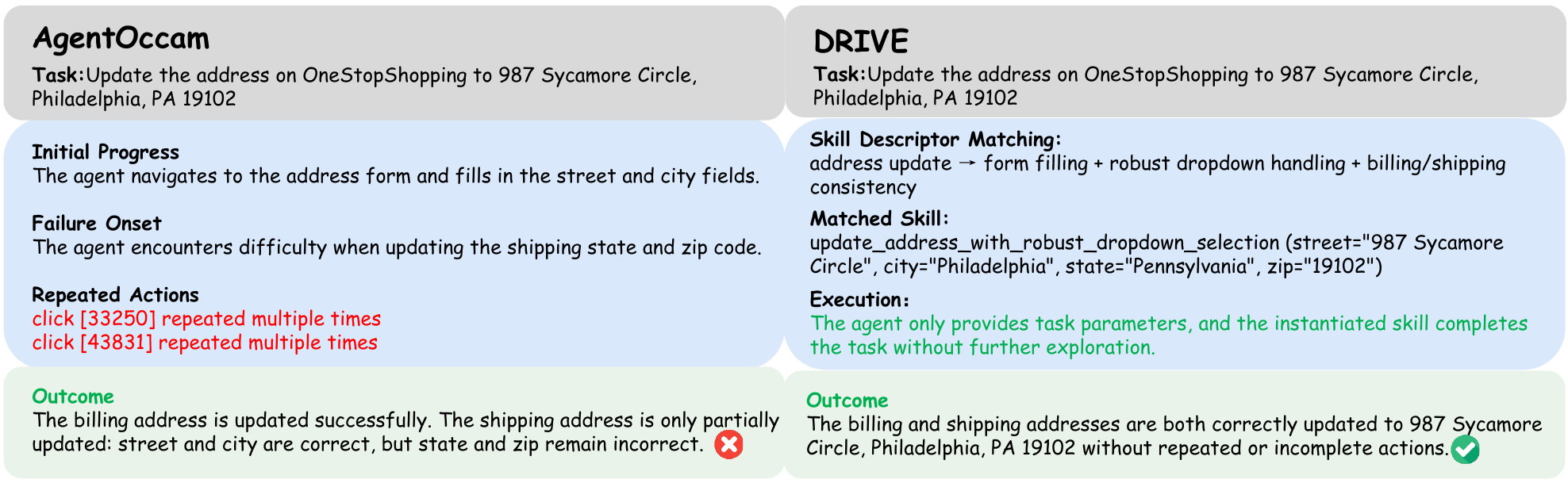}
    \caption{Case study resolving an \textbf{interaction error}. The baseline agent fails due to brittle execution on dynamic interface elements (e.g., state and zip-code field dependencies). In contrast, DRIVE invokes a programmatic interaction skill that encodes a robust procedure for structured form completion, successfully bypassing the page-specific constraints.}
    \label{fig:case1}
\end{figure}

\begin{figure}[h]
    \centering
    \includegraphics[width=\linewidth,trim=0.1cm 0.1cm 0.1cm 0.1cm,clip]{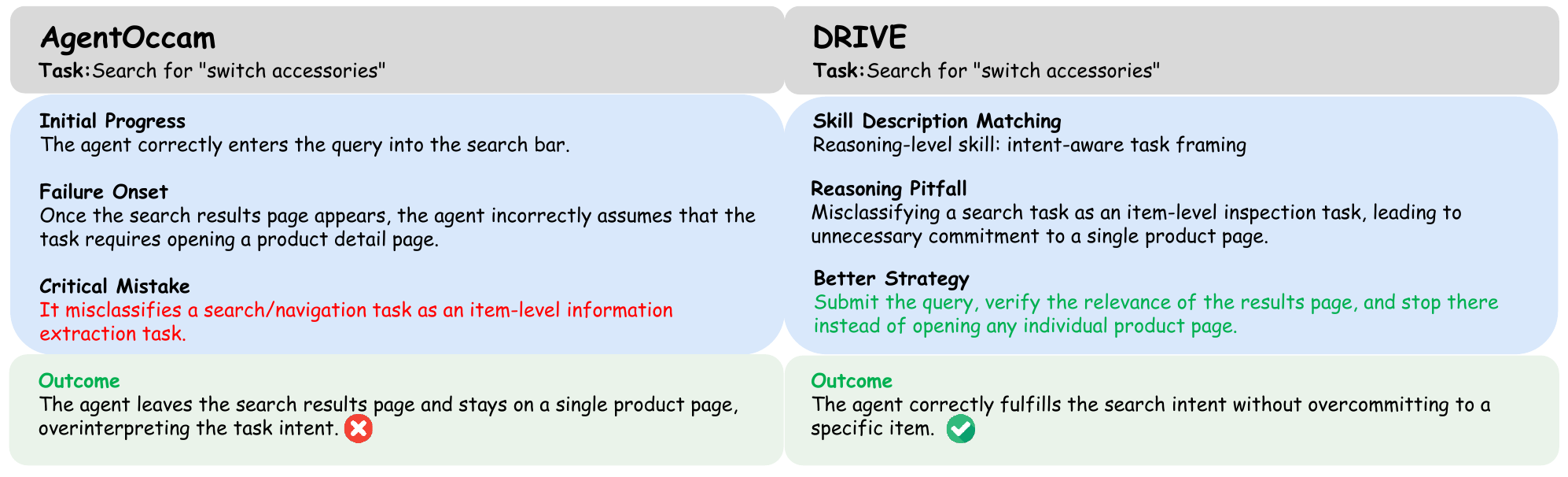}
    \caption{Case study resolving a \textbf{reasoning error}. The baseline agent exhibits flawed stopping-condition judgment, unnecessarily clicking into a product page when the search results already satisfy the query. DRIVE leverages a natural-language reasoning skill to correctly frame the task intent and halt execution at the appropriate state.}
    \label{fig:case2}
\end{figure}

As shown in Fig.~\ref{fig:case1}, the first case reflects a typical interaction-level problem. The agent reaches the target form and fills in several fields correctly, which suggests that it has understood the task intent. However, it repeatedly fails on the state and zip-code fields because success depends on handling field dependencies and dropdown interactions reliably. In this case, additional natural-language guidance is not enough. What is needed is an interaction skill that encodes a robust executable procedure for structured form completion under page constraints.

Fig.~\ref{fig:case2} shows the opposite case. After submitting the query, the agent incorrectly continues to a product page even though the search results page already satisfies the user request. Here, the problem lies not in execution but in task framing and stopping-condition judgment. A more robust interaction procedure would not fix this error, because the agent is pursuing the wrong objective. What is needed instead is a reasoning skill that helps the agent interpret the task correctly and stop at the appropriate point.

Taken together, these two cases show that neither natural-language guidance nor programmatic skills alone are sufficient for web tasks. Effective adaptation requires separate modeling and coordinated use of both reasoning skills and interaction skills.

\subsection{Efficiency and Overhead}
\label{sec:efficiency_overhead}

We further analyze the efficiency overhead introduced by DRIVE. Since skill retrieval and invocation mainly affect the input context, we focus on input-token consumption and interaction cost. As shown in Table~\ref{tab:shopping_overall_eff}, DRIVE improves the Shopping success rate from 37.23\% to 45.99\%, corresponding to an absolute gain of 8.76 percentage points. Although DRIVE increases the average input-token consumption per task by 13.9\%, the input-token cost per successful completion decreases from 164.7k to 151.9k, yielding a 7.8\% reduction. This suggests that the additional context overhead introduced by dual-level skill retrieval is offset by the improved task-completion rate.

\begin{table}[H]
\centering
\small
\setlength{\tabcolsep}{4pt}
\renewcommand{\arraystretch}{1.12}
\begin{tabular*}{\linewidth}{@{\extracolsep{\fill}}lccccc@{}}
\toprule
Method 
& SR (\%) 
& \shortstack{Avg. input\\per task} 
& \shortstack{Input\\per success} 
& \shortstack{Avg. steps\\per task} 
& \shortstack{Avg. calls\\per task} \\
\midrule
Baseline 
& 37.23 
& 61,316 
& 164.7k 
& 9.3 
& 10.8 \\
DRIVE 
& \textbf{45.99} 
& 69,858 
& \textbf{151.9k} 
& \textbf{9.1} 
& 13.3 \\
\midrule
Relative change 
& \textbf{+8.76 pp} 
& +13.9\% 
& \textbf{-7.8\%} 
& \textbf{-2.2\%} 
& +23.2\% \\
\bottomrule
\end{tabular*}
\caption{Overall efficiency and overhead comparison on the Shopping test set ($N=137$). SR denotes task success rate. Input per success is computed by normalizing total input-token consumption by the number of successful completions.}
\label{tab:shopping_overall_eff}
\vspace{-2mm}
\end{table}

\begin{table}[H]
\centering
\small
\setlength{\tabcolsep}{4pt}
\renewcommand{\arraystretch}{1.12}
\begin{tabular*}{\linewidth}{@{\extracolsep{\fill}}lccccc@{}}
\toprule
Method 
& \shortstack{Total input\\tokens} 
& \shortstack{Total\\steps} 
& \shortstack{Avg. input\\per task} 
& \shortstack{Avg. steps\\per task} 
& \shortstack{Input\\per step} \\
\midrule
Baseline 
& 3.35M 
& 464 
& 55.8k 
& 7.7 
& 7.22k \\
DRIVE 
& \textbf{2.98M} 
& \textbf{404} 
& \textbf{49.7k} 
& \textbf{6.7} 
& 7.38k \\
\midrule
Relative change 
& \textbf{-10.9\%} 
& \textbf{-12.9\%} 
& \textbf{-10.9\%} 
& \textbf{-13.0\%} 
& +2.3\% \\
\bottomrule
\end{tabular*}
\caption{Matched-subset efficiency comparison on Shopping. The comparison uses the same matched task subset extracted from the analysis logs, so both methods are evaluated on identical task instances. Token counts are shown in compact form, where M and k denote millions and thousands, respectively. All relative changes are computed with respect to the Baseline.}
\label{tab:shopping_matched_eff}
\vspace{-2mm}
\end{table}

DRIVE also introduces additional orchestration overhead, as reflected by the increase in average calls per task. However, the average number of environment steps slightly decreases from 9.3 to 9.1 on the full Shopping test set. To further examine execution cost under comparable task conditions, Table~\ref{tab:shopping_matched_eff} reports statistics on the same matched task subset extracted from the analysis logs. On this subset, DRIVE reduces total interaction steps by 12.9\%, average steps per task by 13.0\%, and total input-token consumption by 10.9\%. Although the input tokens per step increase slightly, the reduction in trajectory length dominates, leading to lower overall token consumption on the matched subset.

\subsection{Generalization to an Open-Source Backbone}
We also evaluate DRIVE with the smaller open-source backbone Qwen2.5-7B-Instruct~\cite{yang2024qwen25} to test whether its gains extend beyond strong proprietary models. As shown in Table~\ref{tab:webarena_os_sr}, DRIVE improves success rates on all five WebArena domains, with an average relative gain of 52\%. The largest improvement appears on Forum, and the other domains also show clear and consistent gains.

\begin{table}[h]
\centering
\small
\setlength{\tabcolsep}{6pt}
\renewcommand{\arraystretch}{1.1}
\caption{Task success rates on WebArena using Qwen2.5-7B-Instruct. The $\Delta$ row reports relative improvement over the baseline.}
\label{tab:webarena_os_sr}
\begin{tabular}{lcccccc}
\toprule
Method & Shopping & CMS & Forum & Gitlab & Map & Average \\
\midrule
Baseline & 8.4 & 6.1 & 9.9 & 8.7 & 7.8 & 8.2 \\
Ours     & \textbf{12.3} & \textbf{8.9} & \textbf{16.9} & \textbf{12.7} & \textbf{11.7} & \textbf{12.5} \\
\midrule
$\Delta$ & $\uparrow 46\%$ & $\uparrow 46\%$ & $\uparrow 71\%$ & $\uparrow 46\%$ & $\uparrow 50\%$ & $\uparrow 52\%$ \\
\bottomrule
\end{tabular}
\end{table}
Although the absolute success rates remain lower than those achieved with stronger backbones, this is not surprising for long-horizon web tasks. Such tasks require models to handle large, noisy contexts and make reliable multi-step decisions under partial observability. Smaller models are more likely to lose track of relevant state information and make suboptimal decisions during extended interaction, which also reduces the reliability of induced multi-step skills. Even so, the consistent gains in Table~\ref{tab:webarena_os_sr} show that the effectiveness of DRIVE is not tied to a particular high-capacity backbone and can transfer to open-source models as well.

\section{Discussion}
\label{sec:Discussion}

\subsection{Why Dual-Level Skill Modeling Matters}
\label{sec:why_dual_level_skill_modeling_matters}

A key implication of this work is that web-agent experience should not be treated as a single homogeneous memory~\cite{dong2022lifelong}. Web tasks contain two kinds of reusable knowledge: reasoning knowledge, which supports task-level decisions, and interaction knowledge, which grounds those decisions on concrete pages. The former must generalize across tasks, whereas the latter must remain executable under local page constraints. A unified representation is therefore unlikely to preserve both transferability and executability.

This distinction is especially important in dynamic web environments. Interface changes may break an execution procedure while leaving the underlying task strategy valid. Conversely, similar page actions may serve different task goals. Dual-level skill modeling addresses this tension by separating abstract decision knowledge from grounded execution knowledge, allowing the agent to reuse both without forcing them into the same memory form.

\subsection{Continual Improvement Through Skill Maintenance}
\label{sec:continual_improvement_through_skill_maintenance}

Continual improvement requires not only accumulating experience, but also keeping the skill library useful as it grows. More trajectories can expose broader task patterns and page operations, helping the agent adapt to recurring web scenarios. However, this benefit can diminish if outdated or redundant skills impair retrieval.

DRIVE treats skill learning as an iterative maintenance process. New trajectories expand the library, while skill-level feedback revises interaction skills that no longer execute reliably and merges reasoning skills with overlapping guidance. This maintenance reflects the different roles of the two skill types: interaction skills must stay aligned with changing page conditions, whereas reasoning skills must remain sufficiently general for future decisions. In this way, historical experience becomes an evolving capability base rather than a static archive.

\subsection{Limitations and Broader Impacts}
\label{sec:Limitations_and_Broader_Impacts}

Several limitations remain. First, DRIVE depends on reliable failure attribution; incorrect attribution may update the wrong part of the skill library. Second, retrieval may become less precise as the library grows, which calls for more scalable selection mechanisms. Third, our experiments use a DOM-based benchmark and do not fully capture the visual complexity of real websites~\cite{yin2025context}. Extending DRIVE to multimodal web agents is therefore an important direction for future work~\cite{koh2024visualwebarena,zheng2024gpt,he2024webvoyager}.

Continual web-agent learning also raises safety concerns. Agents that improve through experience may become more capable, but they may also reinforce unsafe behaviors if experience is reused without constraints~\cite{qian2026zero}. Future work should combine continual learning with safety checks, sandboxed evaluation, and human oversight for high-risk operations.

\section{Conclusion}

We introduced DRIVE, a dual-level skill modeling framework for continual learning in web agents. DRIVE separates historical experience into transferable reasoning skills and executable interaction skills, and introduces specific mechanisms for scenario-aware retrieval and continuous skill evolution. This design improves both decision making and execution reliability in dynamic web environments. Experiments on five WebArena domains show that DRIVE achieves an average success rate of 52.8\%, outperforming existing baselines. In addition, our ablation studies verify that reasoning skills and interaction skills play distinct but complementary roles in the overall framework. Overall, our results suggest that separating task-level reasoning from page-level execution is an effective way to build web agents that adapt and improve over time.

\section*{Declaration of generative AI and AI-assisted technologies in the manuscript preparation process}
During the preparation of this work, the authors used ChatGPT to improve language expression and refine the cover letter and parts of the manuscript text. After using this tool, the authors carefully reviewed and edited the content as needed and take full responsibility for the content of the published article.

\section*{Acknowledgments}
  This work was supported by the Huxiang Young Talents in Science and Technology Innovation Project (No. 2024RC3148).




\bibliographystyle{elsarticle-num}
\bibliography{cas-refs}
\end{document}